\documentclass[10pt, conference, compsocconf]{IEEEtran}
\usepackage{xcolor}

\usepackage{cite}

\usepackage[pdftex]{graphicx}

\usepackage[cmex10]{amsmath}
\usepackage{ amssymb }
\usepackage{bbold}
\usepackage{multirow, multicol}

\usepackage{algorithmic}

\usepackage{array}
\usepackage[caption=false,font=footnotesize]{subfig}
\usepackage{stfloats}

\usepackage{url}

\begin{document}
%
\title{Real-time instance segmentation with polygons using an Intersection-over-Union loss}


\author{\IEEEauthorblockN{Katia Jodogne-del Litto,  Guillaume-Alexandre Bilodeau}
\IEEEauthorblockA{LITIV Lab., Polytechnique Montréal\\
Montréal, Canada\\
Email: \{katia.jodogne-del-litto, gabilodeau\}@polymtl.ca}
}



\maketitle

\begin{abstract}

Predicting a binary mask for an object is more accurate but also more computationally expensive than a bounding box. Polygonal masks as developed in CenterPoly can be a good compromise. In this paper, we improve over CenterPoly by enhancing the classical regression L1 loss with a novel region-based loss and a novel order loss, as well as with a new training process for the vertices prediction head. Moreover, the previous methods that predict polygonal masks use different coordinate systems, but it is not clear if one is better than another, if we abstract the architecture requirement. We therefore investigate their impact on the prediction. We also use a new evaluation protocol with oracle predictions for the detection head, to further isolate the segmentation process and better compare the polygonal masks with binary masks.  Our instance segmentation method is trained and tested with challenging datasets containing urban scenes, with a high density of road users. Experiments show, in particular, that using a combination of a regression loss and a region-based loss allows significant improvements on the Cityscapes and IDD test set compared to CenterPoly. Moreover the inference stage remains fast enough to reach real-time performance with an average of 0.045 s per frame for 2048$\times$1024 images on a single RTX 2070 GPU. The code is available at: https://github.com/KatiaJDL/CenterPoly-v2.

\end{abstract}


\IEEEpeerreviewmaketitle

\section{Introduction}

The detection and localization of objects of interest are key tasks in computer vision. The demanded accuracy level varies from a rectangle encompassing an object (bounding box), to the production of a binary mask, indicating for each pixel if it is part of the detected object. The latter is the task of instance segmentation, where the objective is to predict a mask for each object of interest while classifying it among predefined categories. These predictions are more accurate than bounding boxes, but also more computationally expensive. Conventional instance segmentation methods can detect the precise location of an object in an image in about 0.2s \cite{he_mask_2017}, which is not fast enough to use in real-time conditions on most hardware. However, to develop individual or collective intelligent vehicles or to improve road safety with traffic monitoring, it is necessary to be able to detect road users in real-time and locate them accurately in a crowded scene. By using mask approximations, the detection speed and therefore the reaction speed for an unexpected event can be greatly increased. 

\begin{figure}[!t]
\centering
\subfloat[Binary mask.]{\includegraphics[width=0.2\textwidth]{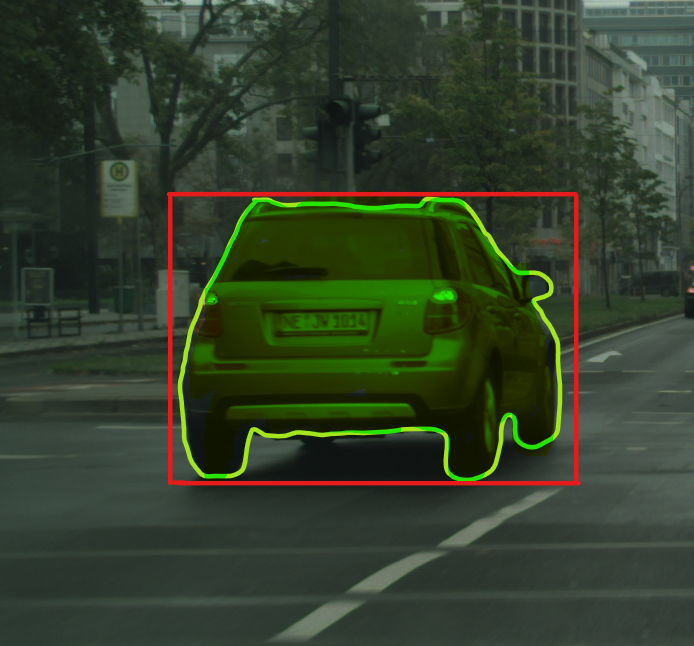}%
\label{fig:binary_gt}}
\hfill
\subfloat[Polygonal mask.]{\includegraphics[width=0.20\textwidth]{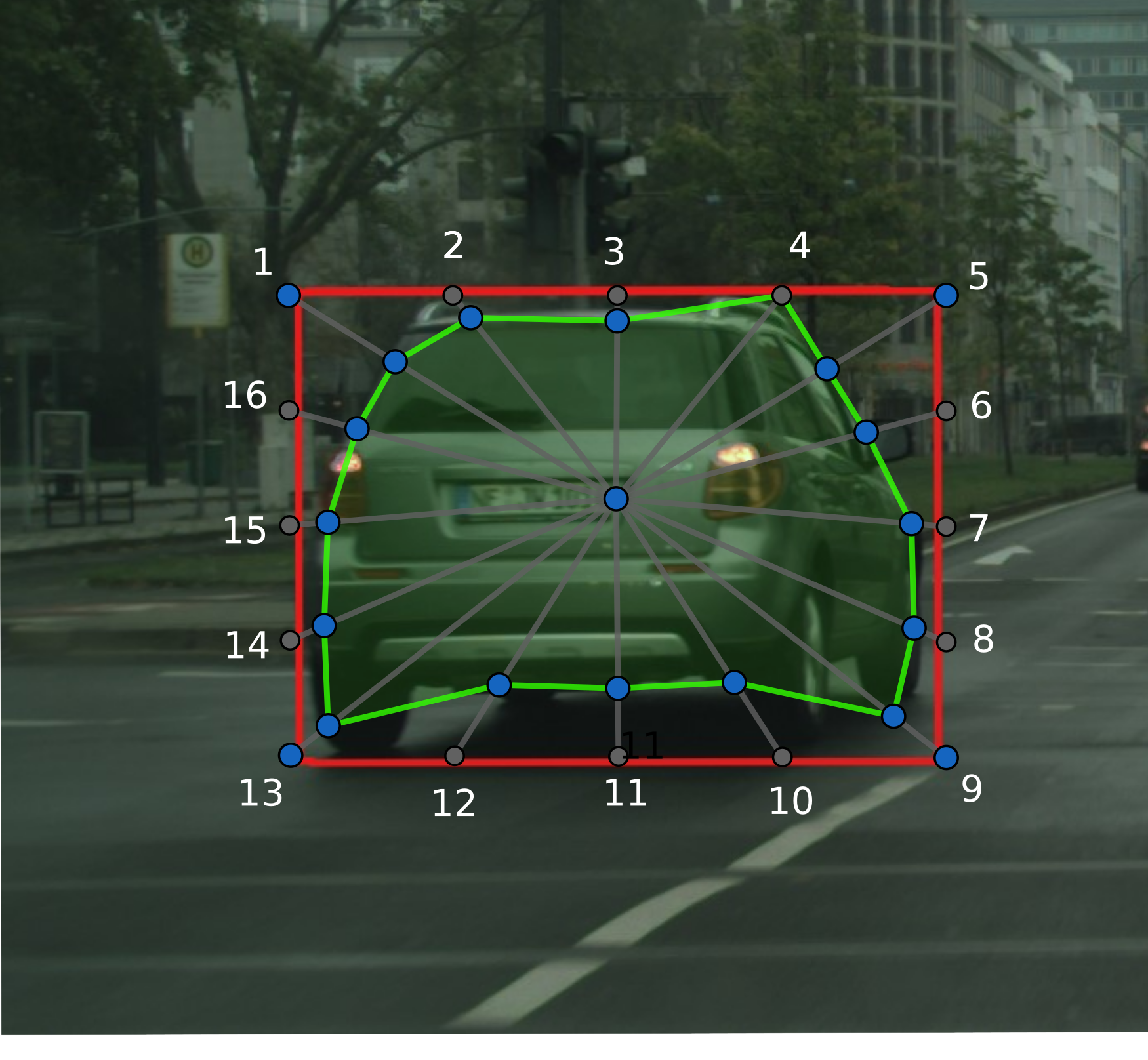}%
\label{fig:poly_gt}}
\caption{Example of ground-truth (GT) binary mask and polygonal mask on a car in Cityscapes dataset \cite{cordts_cityscapes_2016}. Numbers in (b) indicate the casting rays to build the polygon from the GT. Although the polygonal mask does not perfectly match the shape of the object, it can reject a large part of the backgound when compared to the red bounding box.}
\label{fig:gt_masks}
\end{figure}

The intermediate approach we propose builds on CenterPoly by Perreault and al. \cite{perreault_centerpoly_2021}. It consists of using polygonal masks, which are a compromise between bounding box and binary mask. CenterPoly generates simultaneously heatmaps for object detection \cite{zhou_objects_2019} and dense predictions of polygons in the form of vertex sets. The vertices for the ground-truth polygons are produced by casting rays at regular intervals from the bounding box toward its center (Figure \ref{fig:gt_masks}).

To improve over CenterPoly, we investigate the impact of the loss function and the approximation polygon coordinate representation system. 
Our investigation has shown that the precision of the polygonal instance segmentation performed by CenterPoly is not directly limited by the accuracy of the target approximation. As we can see in table \ref{gt_polygons}, the ground-truth polygons created are far more precise than the predictions of CenterPoly.
This motivated us to integrate a region-based loss such as the Intersection-over-Union loss \cite{yu_unitbox_2016} (IoU) to improve the predicted polygons by allowing more flexibility on the vertices position by focusing also on the covered area. No existing loss could be used with vertices coordinates, so we designed a novel polygonal Intersection-over-Union loss using Weiler-Atherton algorithm \cite{weiler_hidden_1977}. Aware of the difficulty to predict relevant coordinates with only a region-based loss, we also propose to add a constraint on the order of the vertices. As no existing loss applied to vertex order, we designed a novel order loss.

\begin{table}[ht]
\renewcommand{\arraystretch}{1.3}
\caption{Ground-truth polygons vs CenterPoly polygons on the Cityscapes validation test.}
\label{gt_polygons}
\centering
\begin{tabular}{llrr}
\hline
Prediction type & Nbr. Vertices  & AP  & AP50\% \\
\hline
 Ground-truth & 16 & 53.0 & 87.7 \\
 Ground-truth &32 & 54.9 & 84.5 \\
\hline
 CenterPoly & 16 & 18.5 & 46.2 \\
 CenterPoly & 32 & 18.4 & 46.0 \\
\hline
\end{tabular}
\end{table}

Moreover, the initial method CenterPoly uses a Cartesian representation of the polygon, whereas other approaches using polygons chose a polar representation \cite{xie_polarmask_2020}, \cite{hurtik_poly-yolo_2022}. In some cases this is necessary for the architecture of the method \cite{xie_polarmask_2020}, in others it is not \cite{hurtik_poly-yolo_2022}, \cite{perreault_centerpoly_2021}, \cite{acuna_efficient_2018}. We therefore studied the impact of the coordinate system on the prediction. 

Finally, to better evaluate the quality of the polygon masks, we propose a new evaluation. The polygon head prediction is attached to  CenterPoly, but can be used independently. To assess precisely the impact of each component, we propose oracle-type experiments to separate polygon prediction and detection. Experiments show that using a Cartesian representation and a combination of a L1 loss and an IoU loss is the best configuration for generating polygonal mask, and allows significant improvements on the Cityscapes and IDD test set compared to CenterPoly. 

Based on these conclusions, we propose a new version of CenterPoly, CenterPolyV2, and our contributions can be summarized as follows:
\begin{itemize}
    \item We present a novel Intersection-over-Union loss function for polygons with Weiler-Atherton algorithm;
    \item We introduce a new loss based on the vertex order;
    \item We study the impact of the geometric representation of polygon vertices for mask approximation, by comparing the Cartesian and polar representations;
    \item We propose a new evaluation experiments to assess more precisely the quality of the generated polygon masks by decoupling detection from the mask prediction. It shows that our method improves significantly over CenterPoly.
\end{itemize}

\section{Related works}

\paragraph{Real-time instance segmentation} Traditional instance segmentation methods, like Mask-RCNN \cite{he_mask_2017}, and its variant PANET \cite{liu_path_2018}, that adds path augmentation, process less than one frame per second. However, few methods can produce masks in real-time. YOLACT \cite{bolya_yolact_2019} accomplishes this by using prototypes and by predicting coefficients to combine them. SOLO \cite{wang_solo_2020} encodes instances as categories, mimicking semantic segmentation methods: The images are divided into cells. Each cell produces a binary mask, corresponding if necessary to the object whose center falls in the cell. SOLOv2 \cite{wang_solov2_2020} is builds upon SOLO and performs further convolution on the features map with a mask kernel predicted separately. ESE SEG \cite{xu_explicit_2019} achieves a speed of 26 ms per frame by performing at the same time the bounding box prediction and segmentation. It uses an approximation of object boundaries based on Chebyshev polynoms. 
SparseInst \cite{cheng_sparse_2022} uses sparse instance activation maps to produce mask kernels but does not need to localize the objects by their center as in SOLO and SOLOv2.
Spatial Sampling Net \cite{mazzini_spatial_2019} is a very fast method which produces a non-uniform density map following object distribution with a diffusion process through a spatial sampling operator. It requires very little post-processing and achieves 113 FPS on Cityscapes. Box2Pix \cite{uhrig_box2pix_2018} combines bounding boxes and semantic segmentation to produce instance masks. It does not quite achieve real-time but it runs on Cityscapes at 10.9 FPS.
Poly-YOLO \cite{hurtik_poly-yolo_2022} and CenterPoly \cite{perreault_centerpoly_2021} both use polygonal mask approximation, and achieve real-time performances on Cityscapes and IDD.

\paragraph{Instance segmentation with masks approximation} A simple way to approximate a mask is to consider only its boundary. 
Deep Snake \cite{peng_deep_2020} uses contour generation and iterative contour deformation to segment instances. Also focusing on contour deformation, PolyTransform \cite{liang_polytransform_2020} converts segmentation masks into polygons to refine the outline. The ExtremeNet method \cite{zhou_bottom-up_2019}, close in principle to CenterNet, predicts the extreme points of the objects, to obtain tight enclosing rectangles. This method allows having easily octagonal masks around the objects. By predicting several points on the outline of the object, it is easy to reconstruct a polygonal mask. However, this technique does not take into account the holes in the objects. Polygon-RNN \cite{castrejon_annotating_2017} uses a recurrent neural network to determine the next vertex, while Polygon-RNN++ \cite{acuna_efficient_2018 } uses also reinforcement learning methods in its training process. These techniques are quite slow and were created for semi-automated annotation creation, not instance segmentation per se.
PolarMask \cite{xie_polarmask_2020} uses a polar representation of the vertices, which are thus at fixed angles and form a star structure. The distance to the center is regressed with an error function based on the Intersection-over-Union. Poly-YOLO \cite{hurtik_poly-yolo_2022}, built on the principle of YOLOv3 \cite{redmon_yolov3_2018}, also uses a polar grid, but the polygons predicted are size-independent and resized during post-processing using the bounding boxes. The number of vertices is dynamic and depends on the object, using a confidence score for each vertex.

Finally, CenterPoly \cite{perreault_centerpoly_2021}, based on CenterNet \cite{zhou_objects_2019}, generates simultaneously heatmaps for object detection and polygon vertices for each pixel. CenterPoly is faster than most of the above-mentioned polygonal instance segmentation methods. It also uses few parameters to represent an objet, without confidence scores.

\section{Method}
\begin{figure*}
    \centering
    \includegraphics[width=0.65\textwidth]{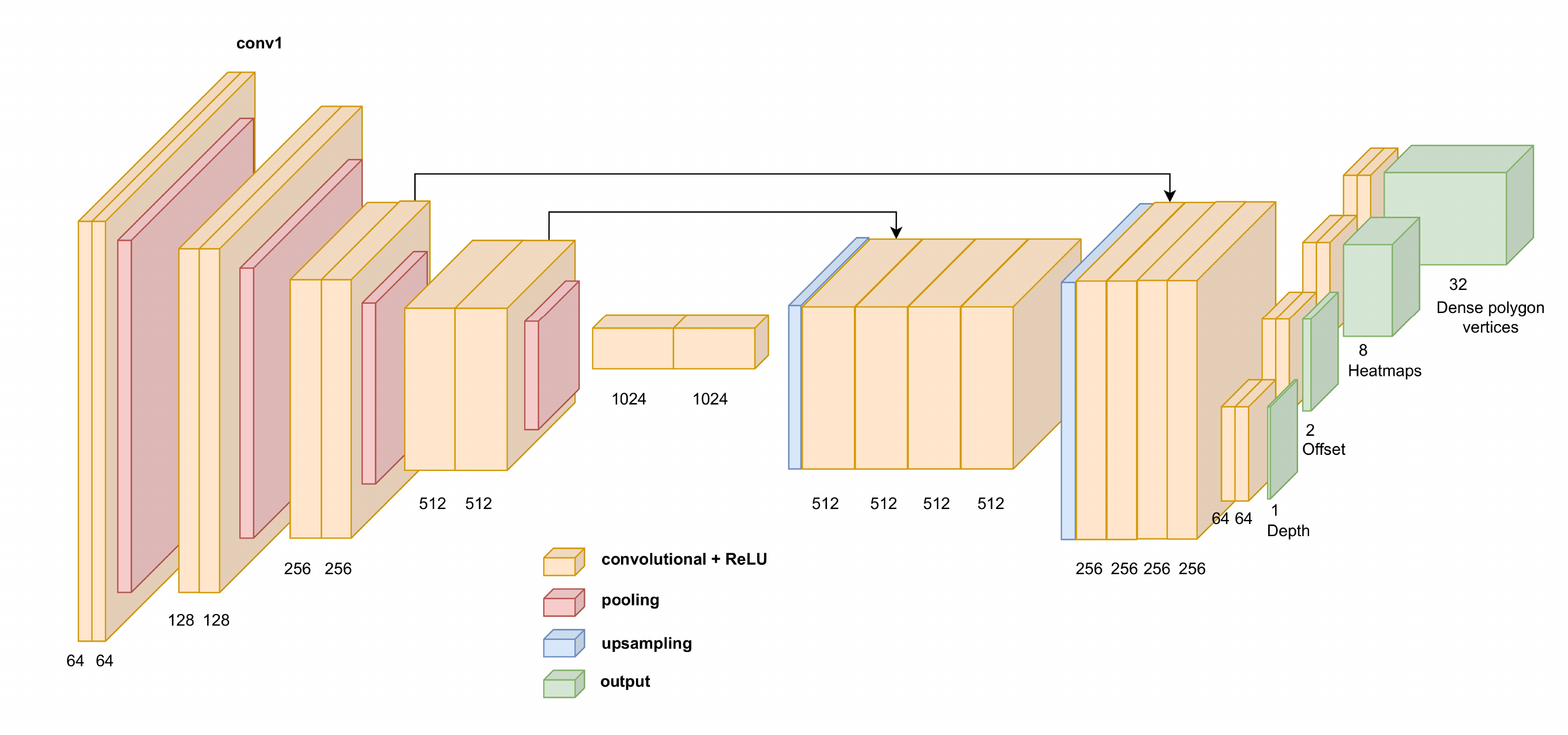}
    \caption{Instance segmentation architecture of CenterPolyV2. The network consists of a CNN backbone, represented here by an hourglass module \cite{newell_stacked_2016}, and four prediction heads: one for heatmaps representing object centers, one for polygon coordinates, one for the offset of the object relative to the center and the last one for the relative depth of the objects. The dimensions are given as a guide and do not reflect exactly the implementation details of the model. See the code for more information.}
\label{fig:architecture}
\end{figure*}

We based our approach on CenterPoly \cite{perreault_centerpoly_2021}, which is built upon the object detector CenterNet \cite{zhou_objects_2019}. The vertices of the polygons are regressed from the center of the objects. Figure \ref{fig:architecture} gives an overview of our CenterPolyV2 network. It is similar to the architecture of CenterPoly. The network consists of a backbone based on a convolutional neural network (CNN), represented here by an hourglass module, and four prediction heads. The generation of heatmaps (one per semantic class) allows to regress the characteristics of the objects from their center. The polygons that are kept being those at the peaks of the heatmaps, corresponding to the centers of the objects of interest. Two more network heads are present. One predicts the offset of the object relative to the center pixel on the heatmap, and the other predicts the relative depth of the objects, which is useful in occlusion cases.

The global loss function of our CenterPolyV2 is given by 
\begin{multline}
    Loss = W_{hm}Loss_{hm} + W_{depth}Loss_{depth} + \\ W_{offset}Loss_{offset} + W_{poly} Loss_{poly}
    \label{eq_global_loss},
\end{multline} 
where $Loss_{hm}$ is a focal loss for the heatmap, $Loss_{depth}$ and $Loss_{offset}$ are L1 losses for the depth prediction and the offset, respectively. These losses are the same as in CenterPoly. The weights $W_{hm}$, $W_{depth}$, $W_{offset}$ and $W_{poly}$ are described in details in section \ref{implementation}.

Our work focuses on the $Loss_{poly}$ term. In CenterPoly, the polygons head performs a regression of polygon vertices, with $Loss_{poly}$ being a L1 loss on the vertices, $Loss_{reg}$.
We investigate the vertices representation and the formulation of the loss function for the polygons head, $Loss_{poly}$. These will be discussed in more detail in the following subsections.

\subsection{Geometric representation of the vertices}

For the vertices representation of the polygons, we can choose between Cartesian and polar coordinates. Since the polygon head generates dense predictions, the coordinates of the vertices are relative to the center from which they are produced. With the Cartesian system, we predict for each vertex the relative distance in terms of height and width ($\Delta x$ and $\Delta y$). In the polar system, we predict the distance to the center ($r$) and the angle from the horizontal axis ($\theta$). The original CenterPoly method uses the Cartesian representation, but this choice is not required by the design. The information predicted by the network in both cases does not have the same geometric meaning. One can therefore wonder if it has an influence on the performance of the method. There is no need for new ground-truth polygons, we can simply do the conversion during pre-processing.

\subsection{Polygonal region-based loss function}

The Intersection-over-Union metric \cite{yu_unitbox_2016} and its variant \cite{rezatofighi_generalized_2019} can serve as a region-based loss function, with some interesting properties. It is closer to our real goal than the L1 regression loss, since it is directly used in the AP metric. The IoU is based on the area covered, and not only on the distance to mask boundaries. Moreover, it treats the instance as a whole and not as $n$ coordinates. Finally, it is scale-independent.

However, defining a IoU loss based on polygons is not trivial. No existing loss could be used for our purpose. Therefore, we designed a novel IoU loss function based on vertices coordinates. First, we need to find the intersection polygon or polygons between the ground-truth and the prediction, and then compute all the necessary areas.  

\begin{figure}
    \centering
    \includegraphics[width=0.7\columnwidth]{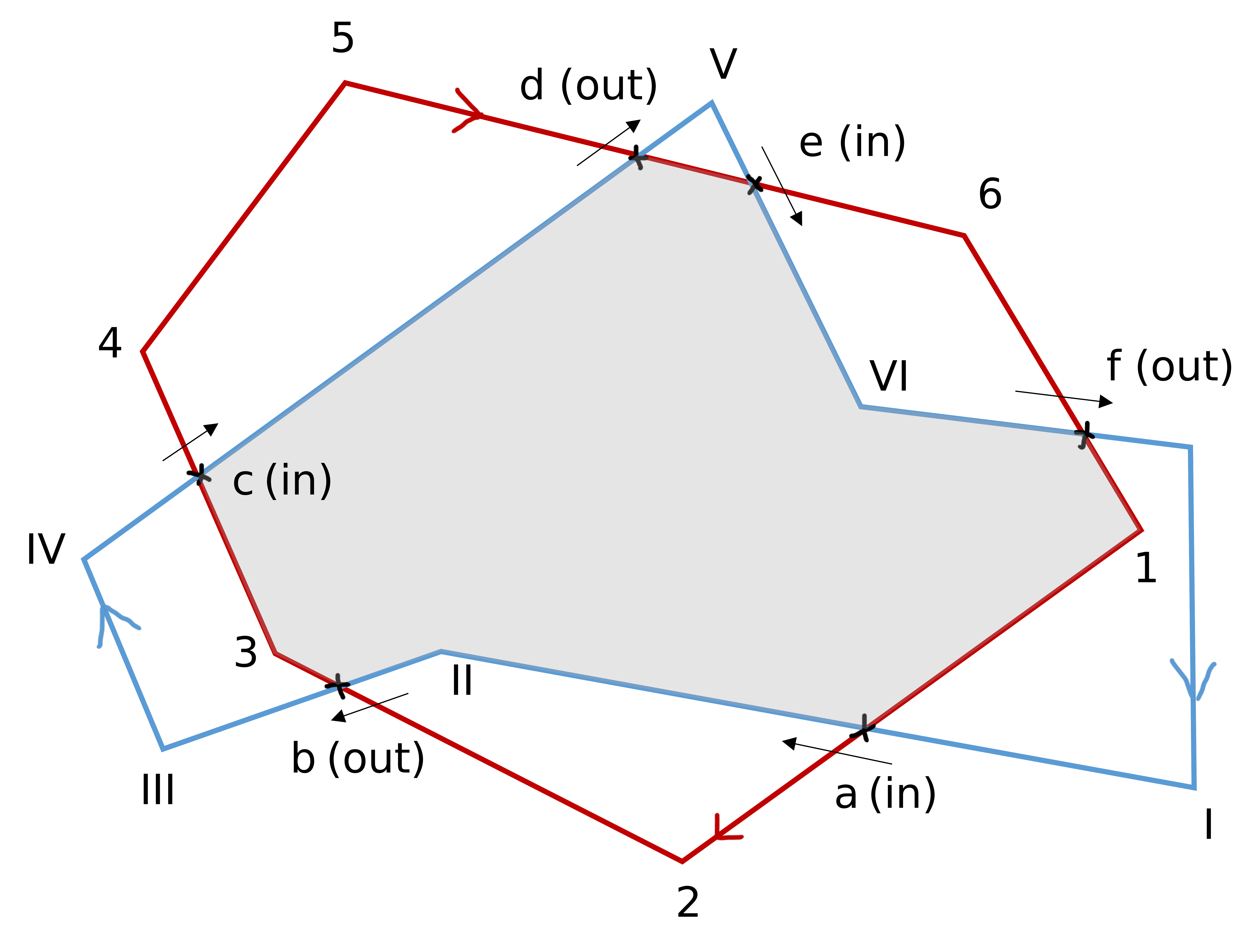}
    \caption{The blue polygon [I, II, III, IV, V, VI] is the predicted polygon (subject polygon, $P_s$), and the red polygon [1, 2, 3, 4, 5, 6] is the ground-truth polygon (clipping polygon, $P_c$). If we start the step 3 of the Weiler-Atherton polygon at the intersection a, the final vertices list of the intersection polygon is [a, II, b, 3, c, d, e, VI, f, 1] (grey area).}
    \label{fig:weiler-atherton}
\end{figure}

We use the Weiler–Atherton algorithm \cite{weiler_hidden_1977} to compute the intersection polygon or polygons. 
Let us define the subject polygon $P_s$ as the predicted polygon, and the clipping polygon $P_c$ as the ground-truth polygon. It can be noted that it is only a convention, they are exchangeable. This algorithm can handle concave polygons, both for polygons $P_c$ and $P_s$. This algorithm allows us to use convex and concave polygons alike, on the condition that they are not self-intersecting. By construction the ground-truth polygons do not intersect themselves, and we ensure the same property for predicted polygons by sorting their vertices along their angles. The algorithm uses polygons represented as a circular list of vertices, in clockwise order, similar to the representation we work with.

The steps of the Weiler–Atherton algorithm are the following (An example of the process is given in figure \ref{fig:weiler-atherton}):
\begin{enumerate}
    \item We compute intersection points between the polygons $P_c$ and $P_s$, using the parametric formulas of the edges. 
    \item We link them to their position in the two vertices lists, with the label "in" (if $P_s$ edge enters the polygon $P_c$) or "out" (if $P_s$ edge exits the polygon $P_c$).
    \item Starting at an "in" intersection, we follow the vertices of $P_s$ until a new intersection is found ("out"). Then we continue on the vertices of $P_c$ until a new intersection is found ("out"), and repeat this step until we find our first intersection.
    \item This makes one intersection polygon. If the "in" intersection list is not empty, we start again at step 3. With concave polygons, the intersection can indeed be composed of multiple polygons.
\end{enumerate}

If the polygons do not intersect (one inside the other, or not overlapping), we consider the area of intersection to be the smallest one of the two. 

To compute the area of the polygons ($P_c$, $P_s$, and intersection polygons) in a differentiable way, we use the shoelace algorithm (or the surveyor formula) \cite{braden_surveyors_1986}, which gives the area of a simple polygon given its vertices' coordinates. The coordinates are assumed to be taken in clockwise order around the polygon, beginning and ending at the same point. The area of a polygon is as follows: \begin{multline}
    \label{eq_area}
    \mathcal{A}_{polygon} = \frac{1}{2}
    \begin{vmatrix}
    x_1 & y_1 \\
    x_2 & y_2 \\
    ... & ... \\
    x_n & y_n \\
    x_1 & y_1
\end{vmatrix} 
    = \frac{1}{2} ((x_1 y_2 + x_2 y_3 + ... + x_1 y_n) \\
     - (x_2 y_1 + x_3 y_2 + ... + x_n y_1)),
\end{multline} 
with $(x_1, y_1), ... (x_n,y_n)$ being the coordinates of the vertices.

With the Weiler-Atherton algorithm and the shoelace algorithm, we have a differentiable way to compute the intersection area of the ground-truth polygon and the predicted polygon. Then  the IoU loss function is given by:

\begin{equation}
    \label{loss_iou}
    Loss_{IoU} = 1 - \frac{\mathcal{A}_{intersection}}{\mathcal{A}_{P_{s}}+\mathcal{A}_{P_{c}}-\mathcal{A}_{intersection}}.
\end{equation}

\subsection{Order-based vertices loss}

Using only a region-based loss like the intersection over union, it is hard to optimize the vertex positions, with the radius and the angle being free in the case of polar coordinates. Polarmask \cite{xie_polarmask_2020} fixes the angles, which simplifies the prediction and the use of the Intersection-over-Union. But this choice reduces the relevance of the ground-truth mask and therefore the accuracy of the prediction. For our case, we designed a constraint based on the order of the vertices to evaluate its influence on vertices prediction. In this case also, we did not find any existing loss that applied to vertex order. Hence, we propose a novel loss.

\begin{figure}
    \centering
    \includegraphics[width=0.6\columnwidth]{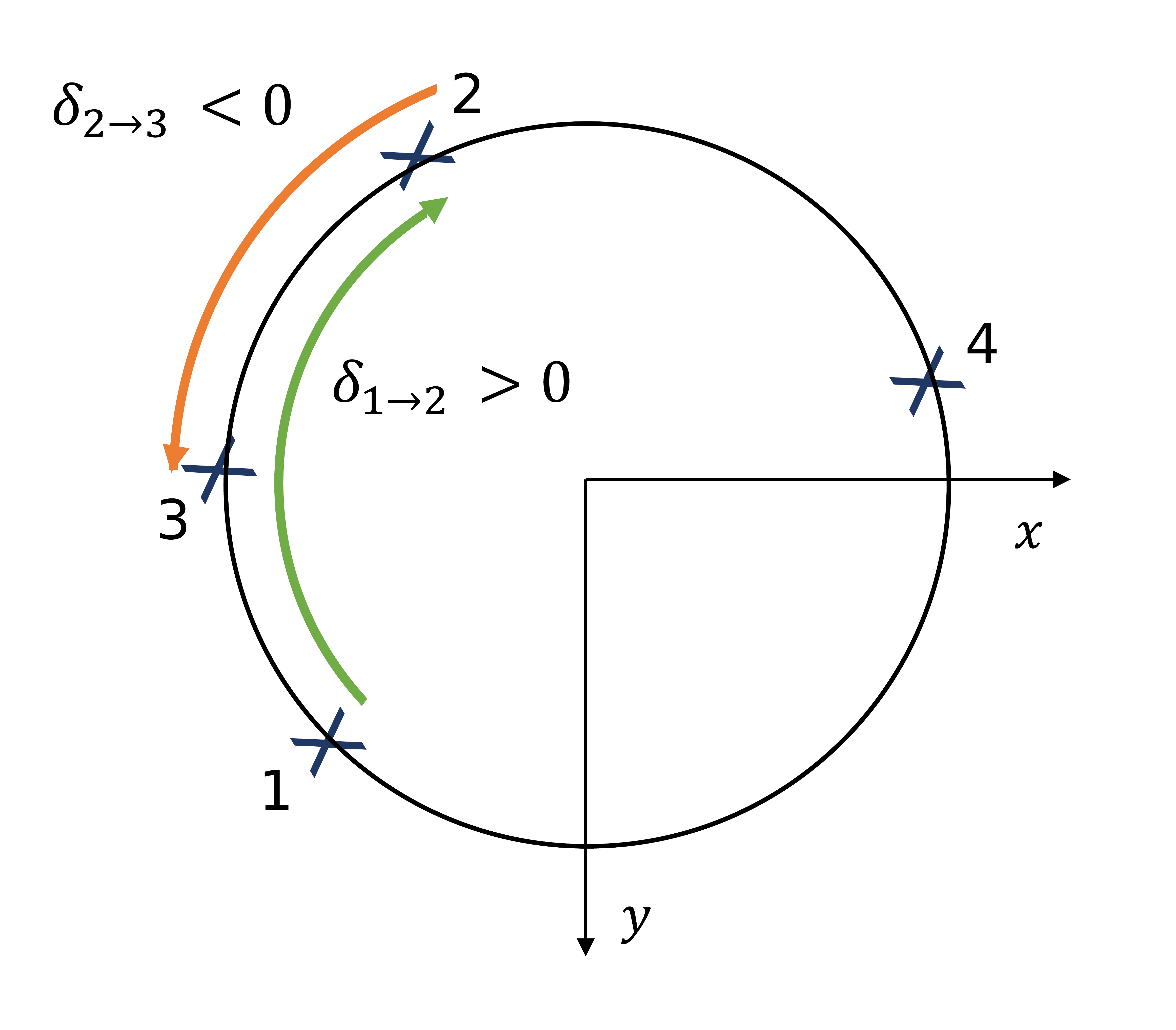}
    \caption{Inversion loss term in the order loss: with these 4 vertices, $Loss_{inversion} = - \delta_{2 \rightarrow 3} = - (\theta_3 - \theta_2) > 0$}
    \label{fig:order_loss}
\end{figure}

This order loss is used when the vertices are represented with polar coordinates. Similar to the ground truth annotations, we wish to predict polygons with clockwise-ordered vertices. It contains two terms. On one hand, we have a constraint on inversion, which sums the differences between inverted angles. It is given by
\begin{equation}
    \label{eq_inversion}
    Loss_{inversion} = \sum_{j=1}^{N-1}{\sum_{i=j+1}^{N}{(\theta_j-\theta_i)\mathbb{1}_{x<0}(\theta_i-\theta_j)}},
\end{equation}
where $\theta_1, ... \theta_N$ are the angles corresponding to the $N$ vertices respectively. Figure \ref{fig:order_loss} shows an example with four vertices and one inversion. 

On the other hand, adding $2\pi$ to an angle does not change its geometrical value. With only a constraint on the inversion, the angles could spread throughout $\mathbb{R}$ and be inverted in a geometrical sense.  The second term prevents this spread and is given by:
\begin{equation}
    \label{eq_spread}
    Loss_{spread} = \sum_{j=1}^{N-1}{\sum_{i=j+1}^{N}{(\theta_j-\theta_i)\mathbb{1}_{x>2\pi}(\theta_i-\theta_j)}}.
\end{equation}

The final order loss function is the combination of these two terms and is given by

\begin{equation}
    \label{eq_orderloss}
    Loss_{order} = Loss_{inversion} + Loss_{spread}.
\end{equation}

\section{Experimental setup}

\subsection{Datasets and evaluation metrics}

\label{implementation}

In our work, we focus on object detection and instance segmentation of road users in dense traffic areas. We trained and evaluated our method on the Cityscapes, IDD, and KITTI datasets, and performed ablation studies and oracle predictions on the Cityscapes validation set.

The Cityscapes dataset \cite{cordts_cityscapes_2016} includes 5,000 densely annotated images recorded in street scenes in Germany. The standard size of the images is $2048 \times 1024$ and we selected the categories of instance corresponding only to road users: car, bicycle, rider, bus, person, motorcycle, truck, train. The Indian Driving Dataset (IDD) \cite{varma_idd_2019} is composed of around 10,000 images of street scenes. The image resolutions are not constant and vary between $1920 \times 1080$ and $1280 \times 964$. The KITTI dataset \cite{geiger_are_2012} for segmentation contains only 200 train images with dense annotations, with image resolution being $1280 \times 384$. The split between training, validation, and testing data is predefined for all datasets.

The evaluation metric for the three datasets is the Average Precision (AP) as defined for the dataset MSCOCO \cite{lin_microsoft_2014}: It is the mean of AP50\% to AP\%95 with steps of 0.05, which represent values of average precision with minimum Intersection over Union from 0.5 to 0.95. The AP50\% is also used. For Cityscapes only, we also have access to two other metrics, AP50m and AP100m, for objects within a range of 50m and 100m respectively. 

\subsection{Implementation details}

We implemented our method with Pytorch \cite{paszke_pytorch_2019} and trained it for 240 epochs on a single RTX 2070 GPU with the adam optimizer \cite{kingma_adam_2015}. Because of its efficiency in CenterPoly, we chose to use the Hourglass network \cite{newell_stacked_2016} with one stack as a backbone for all our experiments. The backbone, heatmap head, and offset head are pre-trained on MSCOCO. We first trained on Cityscapes and then fine-tuned our model for KITTI and IDD. For training, we used classical data augmentation techniques: color augmentation, random cropping, and flipping. The loss weights are $W_{hm} = 1$, $W_{poly} = 1$, $W_{depth} = 0.1$ and $W_{offset} = 0.1$ (Equation \ref{eq_global_loss}). We use a learning rate of 2e-4. For each tested dataset, we divide the learning rate by ten at epochs 90 and 120. As our GPU memory is limited (8 Go), we select a batch size of 4 and a training resolution of $1024 \times 512$.

Following the recommendations of CenterPoly, we use 16 vertices for the best compromise between accuracy and speed. We also kept the elliptical ground truth for the heatmaps and defined the center of each instance as the center of gravity of the vertices.

\section{Results and discussion}

\subsection{Choice of coordinate representation and best loss combination}

\label{section:ablation}

\begin{table}
    \renewcommand{\arraystretch}{1.3}
    \centering
    \caption{Results on the Cityscapes validation set. Rep. stands for vertices representation system. \textbf{Boldface: Best results}.}
    \label{ablation}
    \begin{tabular}{l|lrrr|rr}
    \hline
    Method & Rep. &  L1 & IoU  & order  & AP & AP50\% \\
    \hline
    CenterPoly \cite{perreault_centerpoly_2021} & cartesian & \checkmark & x & x & 20.75 & \textbf{47.20}  \\
    \hline
    CenterPolyV2 & cartesian & \checkmark & \checkmark & x & \textbf{21.46} & 47.16  \\
    CenterPolyV2 & cartesian & x & \checkmark & x & 0.00 & 0.00  \\
    CenterPolyV2 & polar & \checkmark & x & x & 20.15 & 47.08  \\
    CenterPolyV2 & polar & \checkmark & \checkmark & x & 19.46 & 44.84  \\
    CenterPolyV2 & polar & \checkmark & \checkmark & \checkmark & 18.60 & 44.79  \\
    CenterPolyV2 & polar & \checkmark & x & \checkmark &  19.39 & 44.26   \\
    CenterPolyV2 & polar & x & \checkmark & \checkmark &  0.01 & 0.03   \\
    \hline
    \end{tabular}
\end{table}

In table \ref{ablation}, we present results comparing the impact of our contributions: the IoU loss, and the vertices order loss. All components have the same weight in the loss for polygons prediction. We also present a study of the effect of the vertex representation system.

Using the polygonal Intersection-over-Union loss alone is not enough to generate convincing instance segmentation masks. The polygons predicted are not taken into account in the AP metric because the overlap with ground-truth masks does not go over 50\%, the minimal overlap percentage in this computation. The network lacks guidance to place multiple vertices with only global information about the area. But with the help of the $L1$ vertex regression function, the method can produce finer segmentation, especially for the well located masks.

There are no significant differences when using the Cartesian coordinates or polar coordinates. The results with Cartesian coordinates stay slightly better. The polar order loss, combined with the polar representation, reduces slightly the performance. In this particular context, it seems that adding too much constraint to the learning process is counter-productive. 
However, the loss does reduce the number of polygons with self-intersections. Furthermore, even if it did not help as much as we hoped, we believe that this loss could be useful in any applications relying on polar coordinate for training a neural network.

\subsection{Results on test sets}

\label{section_results}

\begin{table*}[t]
\caption{Results on the Cityscapes test set, in the cartesian representation. If the runtimes were not explicitly stated in the original paper, they are estimated based on our knowledge of the method. For the mask types, Full means based on pixel-wise masks,  Polygon means polygonal masks and Outline means boundary-based masks.  Results are taken from the original papers or public online benchmarks, unless stated otherwise. \textbf{Boldface: Best results for real-time methods}. \underline{Underline: Best results overall}.}
  \label{results_cityscapes}
  \centering
  \begin{tabular}{lllrrrrr}
    \hline
    Method & Mask type & Backbone & AP \textuparrow & AP50\% \textuparrow & AP100m \textuparrow & AP50m \textuparrow & Runtime (s) \textdownarrow\\
    \hline
    Mask-RCNN \cite{he_mask_2017} & Full & Resnet-101 & 26.22 & 49.89  & 37.63 & 40.11 & $\simeq$ 0.2 \\
    PANET \cite{liu_path_2018} & Full & FPN & 31.80 & 57.10 & 44.20 & 46.00 & $>$ 1 \\
    PolyTransform \cite{liang_polytransform_2020} & Polygon & Resnet-50-FPN & \underline{40.10} & \underline{65.90} & \underline{54.80} & \underline{58.00} & $>$ 1 \\
    DeepSnake \cite{peng_deep_2020} & Outline & Hourglass-104 & 31.70 & 58.40 & 43.20 & 44.70 & 4.6 \\
    Polygon-RNN++ \cite{acuna_efficient_2018} & Polygon & - & 25.50 & 45.50 & 39.30 & 43.40 & $>$ 1\\ 
    \hline
    Spatial Sampling Net \cite{mazzini_spatial_2019} & Full & - & 9.20 & 16.80	& 16.4 & 21.4 & \textbf{0.009} \\
    Box2Pix \cite{uhrig_box2pix_2018} & Full & GoogLeNet v1 & 13.10 & 27.20 & - & - & 0.092 \\
    Poly-YOLO \cite{hurtik_poly-yolo_2022} & Polygon & SE-Darknet-53 & 11.50 & 26.70 & - & - & 0.049 \\
    Poly-YOLO lite \cite{hurtik_poly-yolo_2022} & Polygon & SE-Darknet-53 & 10.10 & 23.90 & - & - & 0.027 \\
    CenterPoly \cite{perreault_centerpoly_2021} & Polygon & Hourglass-104 & 15.54 & \textbf{39.49} & 23.33 & 24.45 & 0.045 \\
    \hline
    CenterPolyV2 (ours) & Polygon & Hourglass-104 & \textbf{16.64} & 39.42 & \textbf{24.76} & \textbf{27.20}  & 0.045\\
    \hline
  \end{tabular}
\end{table*}

\begin{table}
\caption{Results on the IDD test set, in Cartesian representation. * Results from the original IDD paper \cite{varma_idd_2019}. \textbf{Boldface: Best results for real-time methods}. \underline{Underline: Best results overall}.}
  \label{results_idd}
  \centering
  \begin{tabular}{lllrrr}
    \hline
    Method & Backbone & AP  & AP50\% & Time (s) \\
    \hline
    Mask-RCNN \cite{he_mask_2017}* & Resnet-101 & 26.80 & 49.90 & $\simeq$ 0.2 \\
    PANET \cite{liu_path_2018}*& FPN & \underline{37.60} & \underline{66.10} & $>$ 1 \\
    \hline
    Poly-YOLO \cite{hurtik_poly-yolo_2022} & SE-Darknet-53 & 11.50 & 26.70 & 0.049 \\
    Poly-YOLO lite \cite{hurtik_poly-yolo_2022}  & SE-Darknet-53 & 10.10 & 23.90 & \textbf{0.027} \\
    CenterPoly \cite{perreault_centerpoly_2021}  & Hourglass-104 & 14.40 & 36.90 & 0.045 \\
    \hline
    CenterPolyV2 (ours)  & Hourglass-104 & \textbf{17.40} & \textbf{45.10} & 0.045  \\
    \hline
  \end{tabular}
  
\end{table}

\begin{table}
\caption{Results on the KITTI test set, in the Cartesian representation. \textbf{Boldface: Best results}}.
  \label{results_kitti}
  \centering
  \begin{tabular}{lllrrr}
    \hline
    Method &  Backbone & AP & AP50\% & Time (s)\\
    \hline
    CenterPoly \cite{perreault_centerpoly_2021}  & Hourglass-104 & 8.73 & 26.74 & \textbf{0.045} \\
    \hline
    CenterPolyV2 (ours)  & Hourglass-104 & \textbf{8.86} & \textbf{26.86} & \textbf{0.045} \\
    \hline
  \end{tabular}
  
\end{table}

Given the results of our study about the loss and coordinate system representation, our final method, CenterPolyV2, includes both the L1 loss and polygonal IoU loss, with a Cartesian vertices representation. The loss function for the polygons head is as follows:
\begin{equation}
Loss_{poly} = Loss_{reg}   + Loss_{IoU},
\end{equation}
with $Loss_{IoU}$ corresponding to Equation \ref{loss_iou}.

We present our results on the three datasets in the table \ref{results_cityscapes} for Cityscapes, in table \ref{results_idd} for IDD, and in table \ref{results_kitti} for KITTI. For comparison purpose, we include slower and more precise methods.

For the Cityscapes test set, CenterPolyV2 improves CenterPoly by 1.1 AP, and the two AP with distance constraints by 1.4 and 2.8 for respectively AP100m and AP50m, but the AP50\% metric stays the same. So the Intersection-over-Union loss function improves the segmentation mask when the representation is already good (with IoU superior to 50\%). The vertices must be already well predicted, so that the IoU loss can optimize their position by taking into account the area. 

The Indian Driving Dataset is more challenging, with more variety in terms of scene settings. CenterPolyV2 reaches 17.40 AP and 45.10 AP50\%. With the combination of L1 and IoU loss functions, the accuracy of the predicted polygons is improved. The IoU loss is indeed less sensitive to the inter-categorical diversity because it takes into account the whole object and not only the distance from the center to the outline. For the KITTI dataset, there is no significant increase in the accuracy, be it for the AP or AP50. This stagnation may be related to the small size of the dataset.

Overall, adding the Intersection-over-Union loss function to the L1 regression loss for the polygon brings improvement to the masks that were already well predicted, but not well adjusted on the object. As other fast methods, we do not yet approach the performance of the best networks that do not consider the speed of inference as a priority. 

The runtime results given in table \ref{results_cityscapes} are taken from the original papers. Some popular methods have since been re-implemented more efficiently. To ensure the relevance of developing new models, we measured the execution time of Mask-RCNN in its recent implementation with Detectron 2 \cite{wu_detectron2_2019} with the same hardware as CenterPolyV2 (see section \ref{implementation}). Results observed in Table \ref{results_runtime} show that these traditional methods remain slow on less powerful infrastructure. Compared to other fast methods, ours ranks midway in runtime, but first in AP, showing a good compromise between speed and mask quality. 

\begin{table}
\caption{Average inference time over the test set of Cityscapes. All inference times are taken on the same computer on a single RTX2070. * Detectron2 implementation \cite{wu_detectron2_2019}}
  \label{results_runtime}
  \centering
  \begin{tabular}{llrrr}
    \hline
    Method & backbone & Runtime(s)  \\
    \hline
    Mask-RCNN \cite{he_mask_2017}* & Resnet-50-FPN & 0.3 \\
    CenterPoly \cite{perreault_centerpoly_2021}& Hourglass-104 & 0.045 \\
    CenterPolyV2 & Hourglass-104 & 0.045 \\
    \hline
  \end{tabular}
\end{table}

\begin{figure*}
    \centering
    \subfloat{\includegraphics[width=0.3\textwidth]{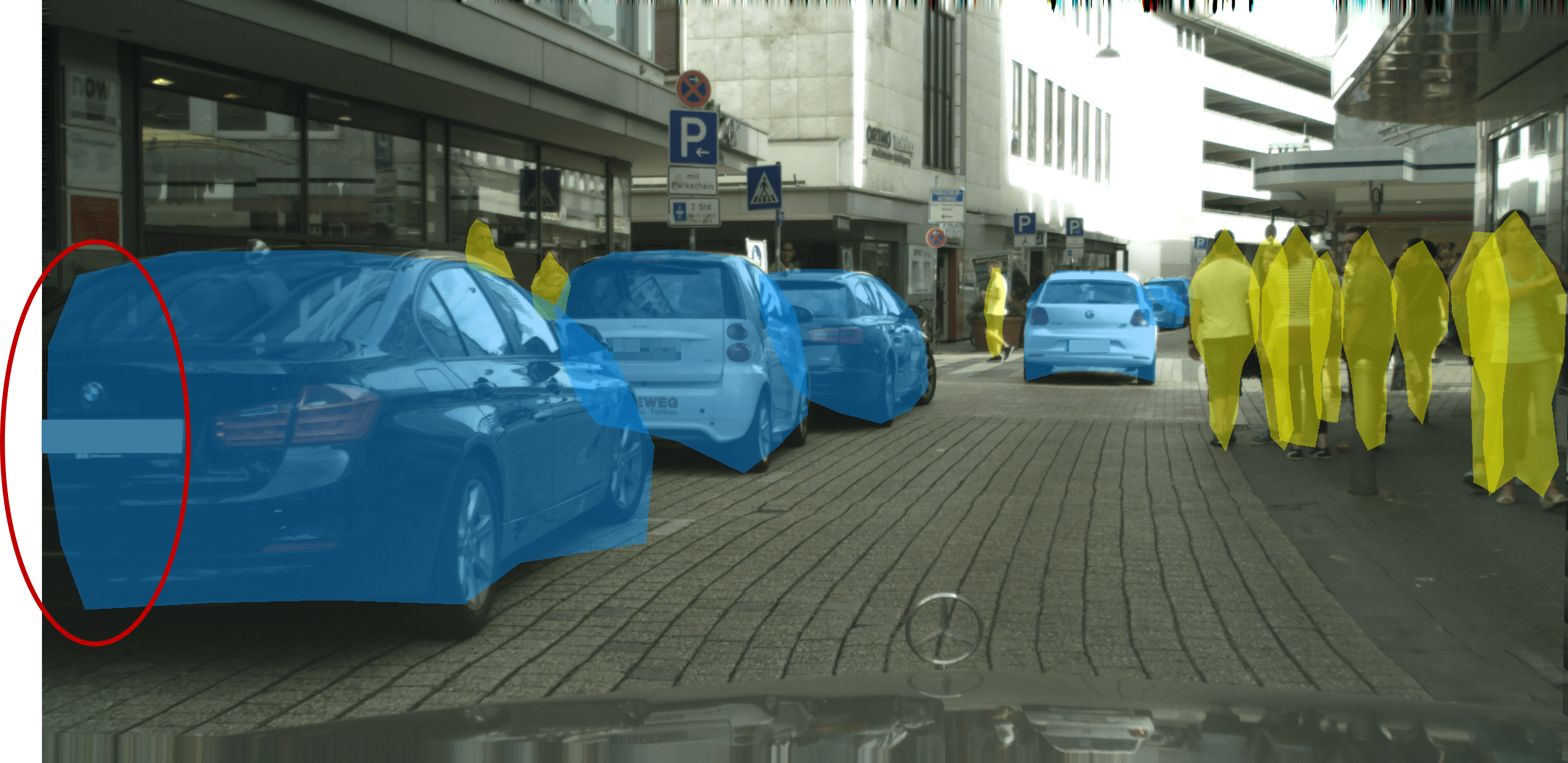}}
    \hfill
    \subfloat{\includegraphics[width=0.29\textwidth]{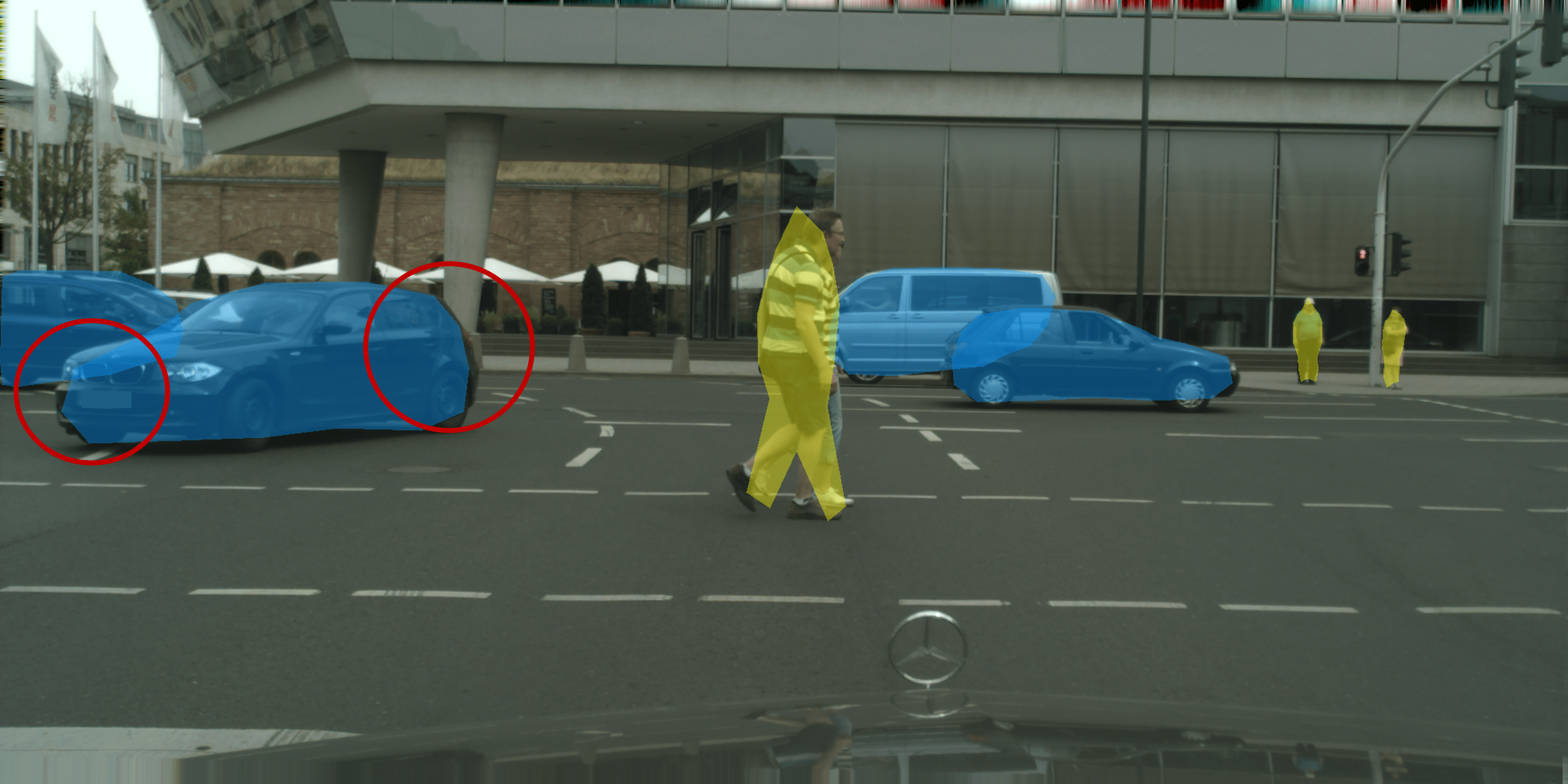}}
    \hfill
    \subfloat{\includegraphics[width=0.29\textwidth]{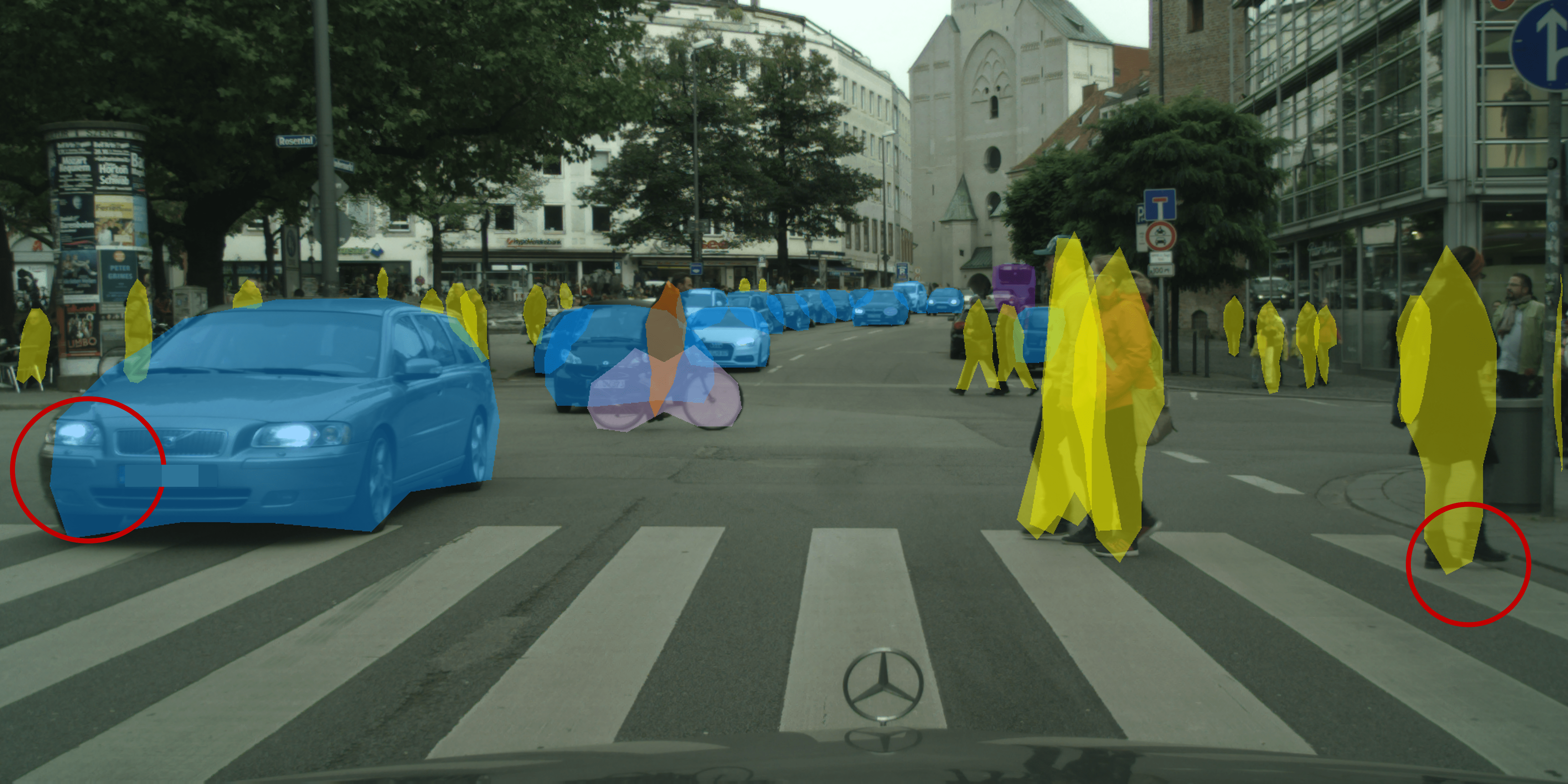}}
    \hfill
    \subfloat{\includegraphics[width=0.3\textwidth]{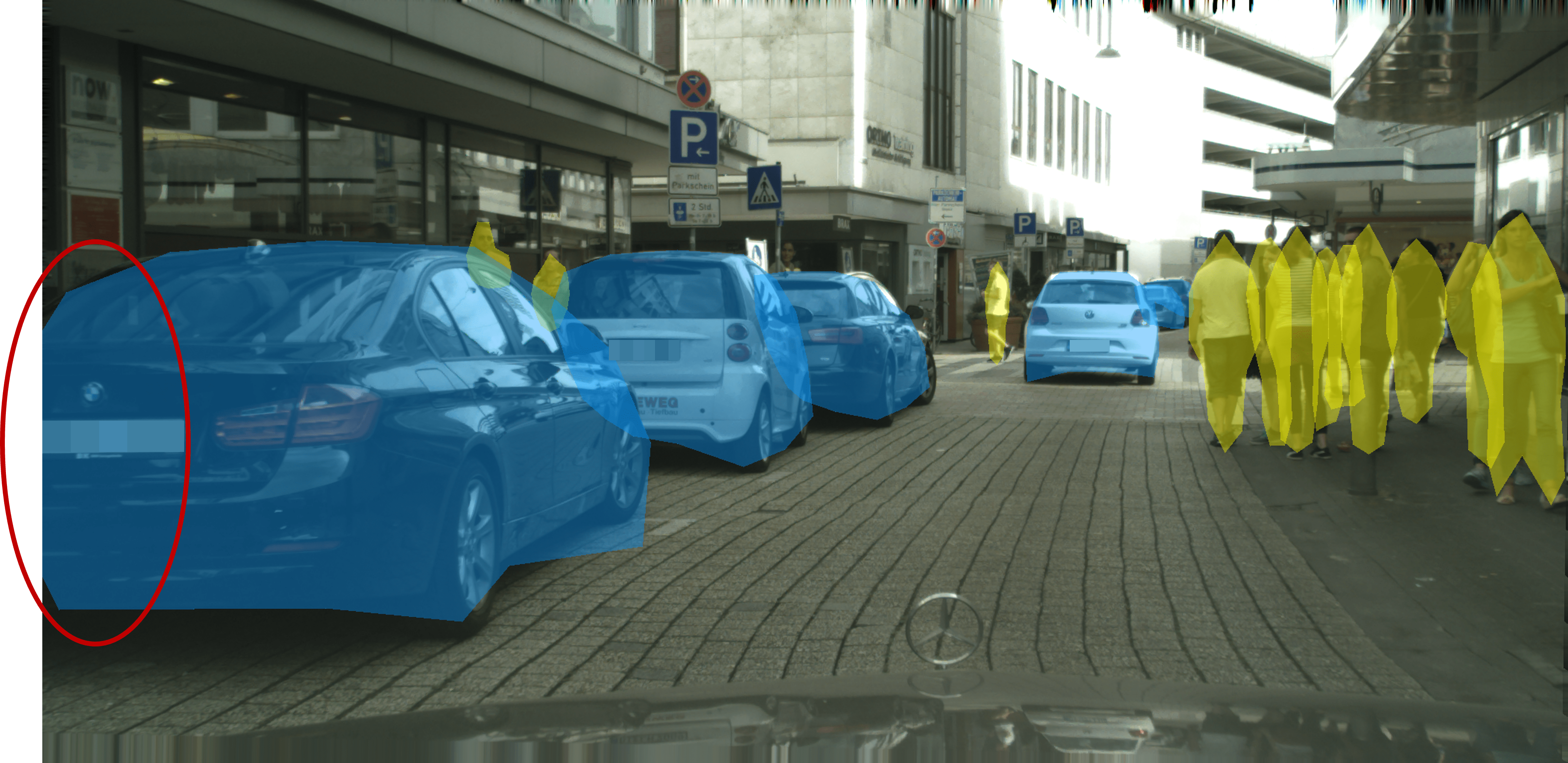}}
    \hfill
    \subfloat{\includegraphics[width=0.29\textwidth]{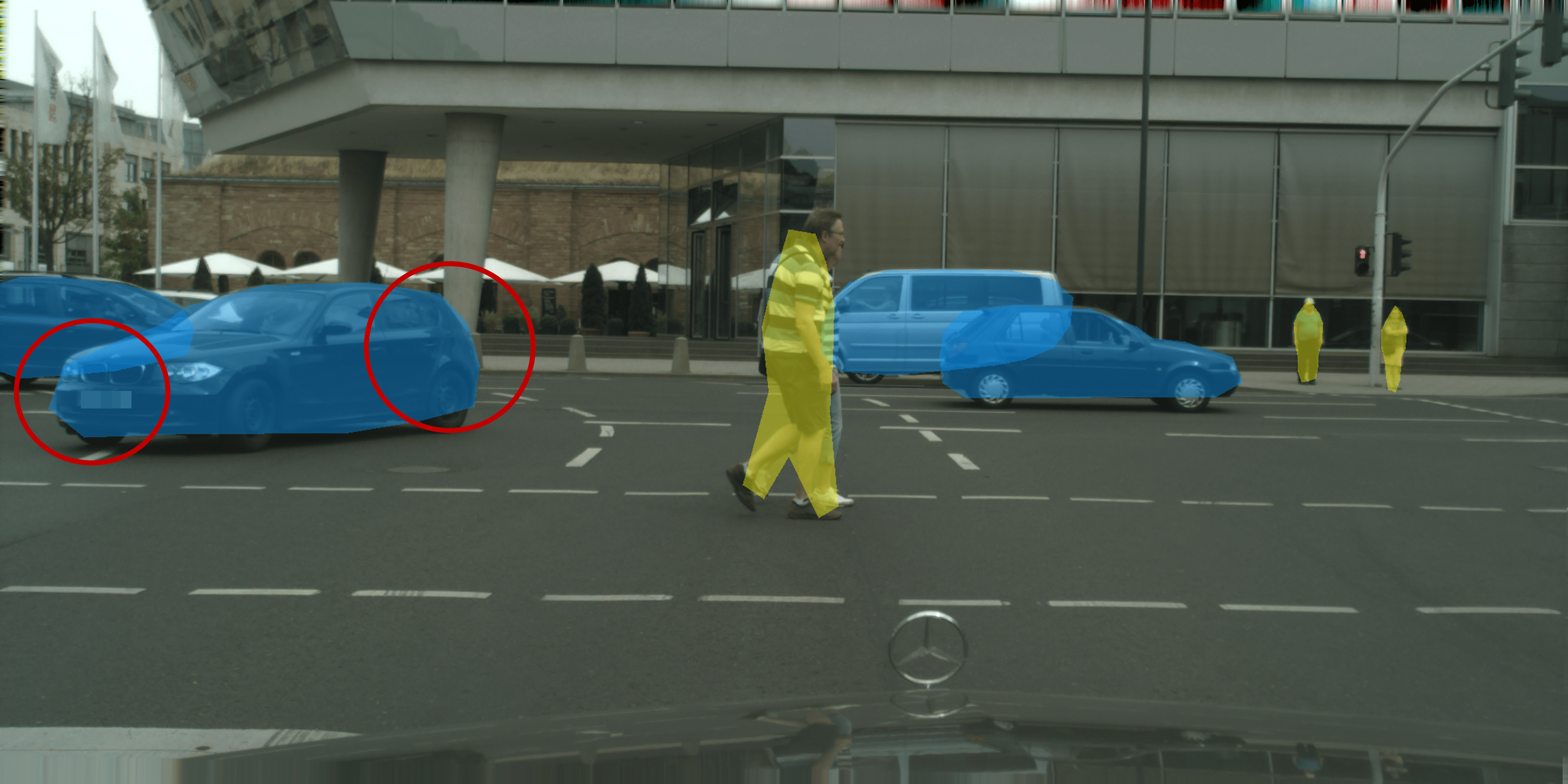}}
    \hfill
    \subfloat{\includegraphics[width=0.29\textwidth]{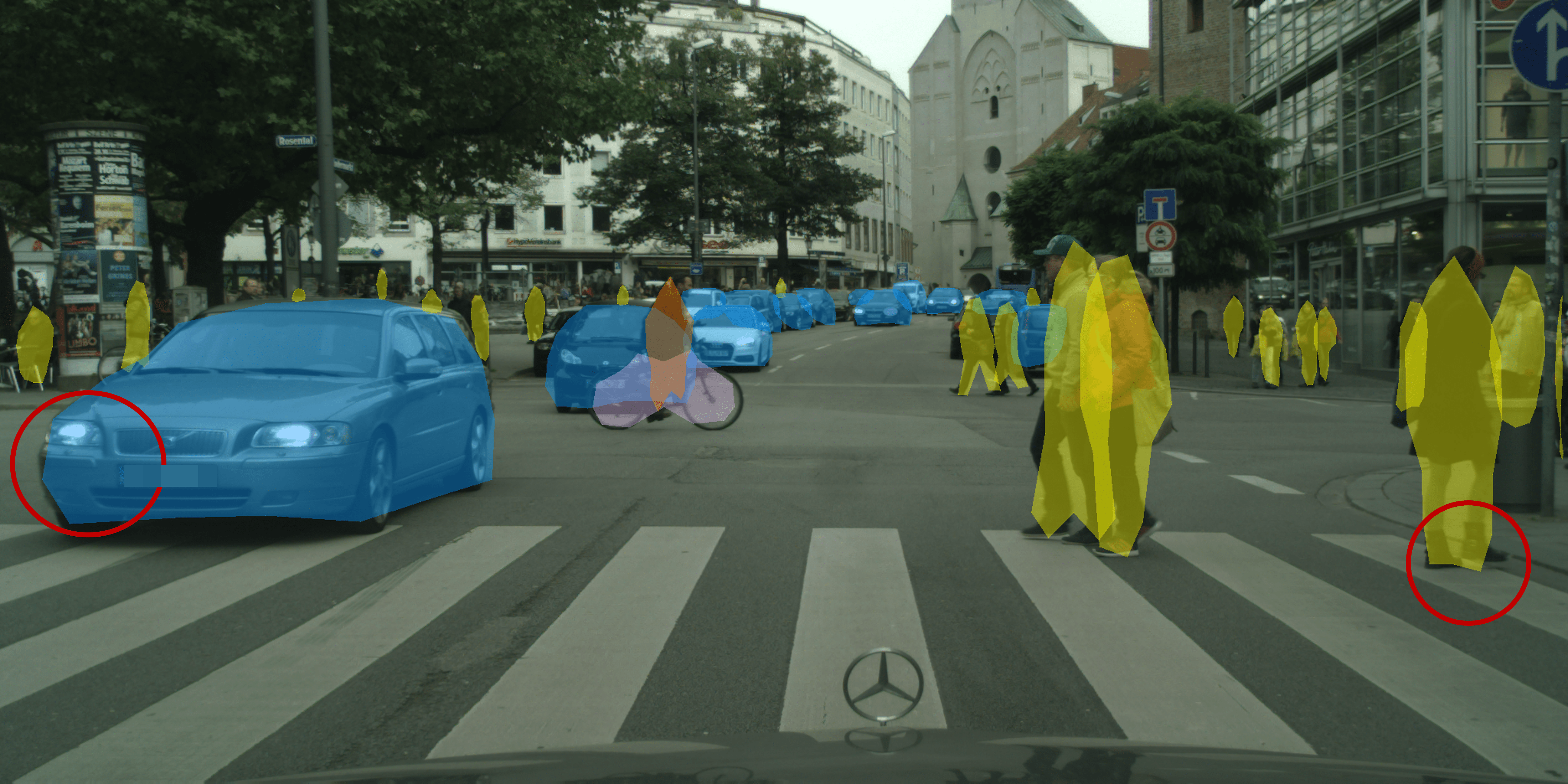}}
    \caption{Qualitative results on the Cityscapes test set. Row 1: CenterPoly. Row 2: CenterPolyV2. Colors correspond to semantic categories. Red circles indicate visible differences.}
    \label{fig:qualitative_results}
\end{figure*}

Qualitative results shown in figure \ref{fig:qualitative_results} support our interpretation. When zooming in, we can see that the mask coverage is improved for objects that were already well predicted in CenterPoly, like cars and some of the pedestrians.

\subsection{Oracle predictions}

Our contributions are not directly related to the architecture of the whole instance segmentation method. We employ the CenteNet detection method as a basis for its lightness and its speed. However, the polygon prediction head could be used with another detection network. Therefore, we decoupled the detection and segmentation tasks to better show our contribution. On the validation set of Cityscapes, we replace the predicted heatmaps by the ground-truth heatmaps and select the predictions from the three other heads according to the ground-truth centers. It simulates the behavior of the polygon head with a "perfect" detection head. Results are presented in table \ref{ablation_oracle}. The accuracy increases notably and the differences we noticed in section \ref{section:ablation} are accentuated. When assuming perfect detection, we find a 6.5 AP difference between CenterPoly and CenterPolyV2. This shows that performance benefits greatly from the addition of our polygonal IoU loss. However, the performance still does not reach the accuracy of the ground-truth polygons (Table \ref{gt_polygons}).

\begin{table}
    \renewcommand{\arraystretch}{1.3}
    \centering
    \caption{Results on the Cityscapes validation set. Oracle predictions are used for the heatmap head. Rep. stands for vertex representation system. \textbf{Boldface: Best results}.}
    \label{ablation_oracle}
    \begin{tabular}{l|lrrr|rr}
    \hline
    Method & Rep. &  L1  & IoU  & order  & AP & AP50\% \\
    \hline
    CenterPoly & cartesian & \checkmark & x & x & 23.46 &  \textbf{54.38} \\
    \hline
    CenterPolyV2 & cartesian & \checkmark & \checkmark & x & \textbf{29.98} & 54.09  \\
    CenterPolyV2 & cartesian & x & \checkmark & x & 0.01 & 0.03  \\
    CenterPolyV2 & polar & \checkmark & x & x & 21.34 & 52.76  \\
    CenterPolyV2 & polar & \checkmark & \checkmark & x & 21.10 & 52.11  \\
    CenterPolyV2 & polar & \checkmark & \checkmark & \checkmark & 20.21 &  50.71 \\
    CenterPolyV2 & polar & \checkmark & x & \checkmark &  20.59 & 50.78   \\
    CenterPolyV2 & polar & x & \checkmark & \checkmark & 0.01  &  0.05  \\
    \hline
    \end{tabular}
\end{table}

\section{Conclusion}

In this paper, we show that we can improve polygonal instance segmentation with an Intersection-over-Union loss function. Combined with L1 loss, the polygonal IoU loss improves the global accuracy for instance segmentation in dense urban scenes, especially for the objects that are already well predicted by not accurately segmented. However, there are inherent limitations to the choice of polygonal approximations. One other possible approach could be to change the approximation type: Instead of predicting the outline, we could find a way to represent directly the inside of the object with only a few parameters.

\section*{Acknowledgment}

We acknowledge the support of the Natural Sciences and Engineering Research Council of Canada (NSERC), the Institut for data valorisation (IVADO) and the Apogée fund.





\def\url#1{}
\bibliographystyle{IEEEtran}
\bibliography{IEEEabrv,Poly}
%



\end{document}